\documentclass{article}
\usepackage{ijcai19}

\usepackage{times}
\usepackage{soul}
\usepackage{url}
\usepackage[hidelinks]{hyperref}
\usepackage[utf8]{inputenc}
\usepackage[small]{caption}
\usepackage{graphicx}
\usepackage{amsmath}
\usepackage{booktabs}
\usepackage{epsfig}
\usepackage{epstopdf}
\usepackage{makecell,multirow,diagbox}
\usepackage{color}
\title{Omnidirectional Scene Text Detection with Sequential-free Box Discretization}

\author{
Yuliang Liu$^1$\and
Sheng Zhang$^1$\and
Lianwen Jin$^1$\footnote{Corresponding author: Lianwen Jin.}\and
Lele Xie$^1$\and
Yaqiang Wu$^2$ \textnormal{and} Zhepeng Wang$^2$
\affiliations
$^1$School of Electronic and Information Engineering, South China University of Technology, China \\
$^2$Lenovo Inc, China\\
\emails
liu.yuliang@mail.scut.edu.cn; lianwen.jin@gmail.com
}
\begin{document}

\maketitle

\begin{abstract}
Scene text in the wild is commonly presented with high variant characteristics. Using quadrilateral bounding box to localize the text instance is nearly indispensable for detection methods. However, recent researches reveal that introducing quadrilateral bounding box for scene text detection will bring a label confusion issue which is easily overlooked, and this issue may significantly undermine the detection performance. 
To address this issue, in this paper, we propose a novel method called Sequential-free Box Discretization (SBD) by discretizing the bounding box into key edges (KE) which can further derive more effective methods to improve detection performance. 
Experiments showed that the proposed method can outperform state-of-the-art methods in many popular scene text benchmarks, including ICDAR 2015, MLT, and MSRA-TD500. Ablation study also showed that simply integrating the SBD into Mask R-CNN framework, the detection performance can be substantially improved.
Furthermore, an experiment on the general object dataset HRSC2016 (multi-oriented ships) showed that our method can outperform recent state-of-the-art methods by a large margin, demonstrating its powerful generalization ability. Source code: {\color{blue} \url{https://github.com/Yuliang-Liu/Box\_Discretization\_Network}}.

\end{abstract}

\section{Introduction}\label{sec:intro}
 Scene text presented in real images are often found with multi-oriented, low quality, perspective distortions, and various sizes or scales. To recognize the text content, it is an important prerequisite for detecting methods to localize the scene text tightly.

\begin{figure}[htb]
\begin{minipage}[c]{0.49\linewidth}
  \centering
  \centerline{\includegraphics[width = 4.3cm, height = 4.3cm]{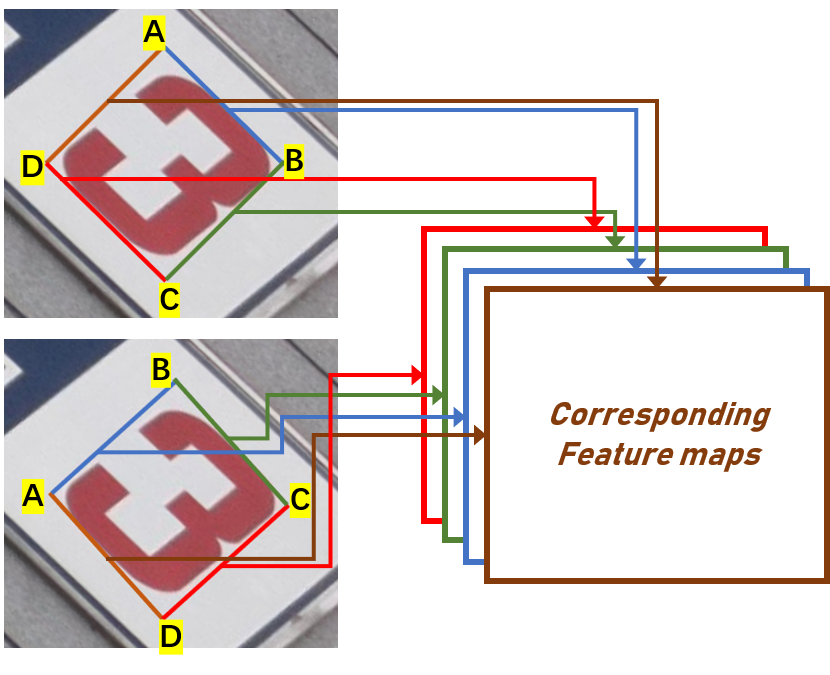}}
  \centerline{\small{(a) Sensitive to label sequence. }}\medskip
\end{minipage}
\hfill  
\begin{minipage}[c]{0.49\linewidth}
  \centering
  \centerline{\includegraphics[width = 4.3cm, height = 4.3cm]{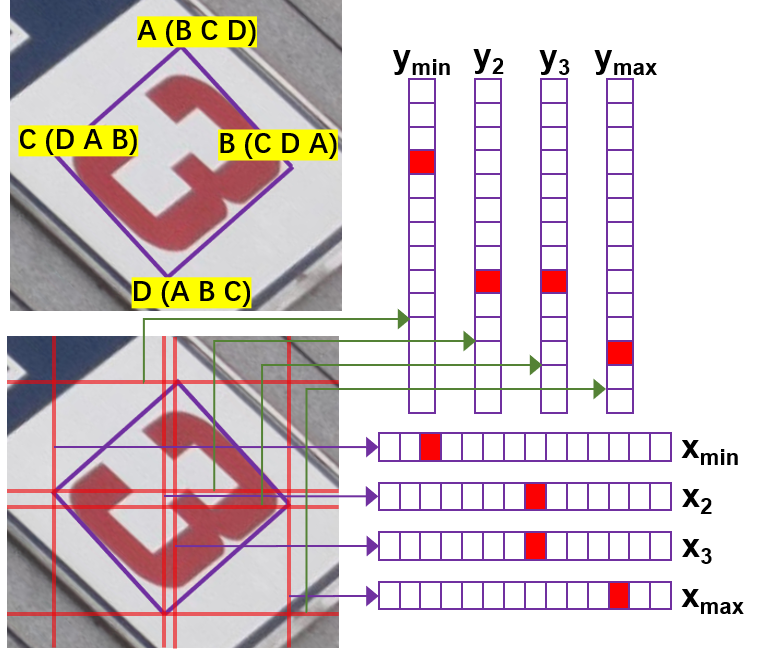}}
  \centerline{\small{(b) Irrelevant to label sequence.  }}\medskip
\end{minipage}

\caption{(a) Previous detecting methods that are sensitive to the label sequence. (b) The proposed SBD.}\label{fig:intro_2}
\end{figure}
 Recently, scene text detection methods have achieved significant progress \cite{zhou2017east,liu2017deep,deng2018pixellink,liao2018textboxes++}. One reason for the improvement is that these methods introduce rotated rectangles or quadrangles instead of axis-aligned rectangles to localize the oriented instances, which remarkably improves the detection performance. However, performance of current methods still have a large gap to bridge a commercial application. Recent studies \cite{liu2017deep,zhu2018sliding} have found that an underlying problem of introducing quadrilateral bounding box may significantly undermine the detection performance.

 Taking East \cite{zhou2017east} as an example: For each pixel of the high-dimensional representation, the method utilizes four feature maps corresponding to the distances from this pixel to the ground truth (GT). It requires preprocessing steps to sort the label sequence of each quadrilateral GT box so that each predicted feature map can well focus on the targets, otherwise the detecting performance may be significantly worse. Such method is called ``Sensitive to Label Sequence'' (SLS), as shown in Figure \ref{fig:intro_2} (a). The question is that it is not trivial to find a proper sorting rule that can avoid Learning Confusion (LC) caused by sequence of the points. The rules proposed by \cite{liu2017deep,liao2018textboxes++,he2018end} can alleviate the problem; however, they cannot avoid that a single pixel deviation of a man-made annotation may totally change the corresponding relationships between each feature map and each target of the GT. 

\begin{figure*}[htb]
  \centering
  \centerline{\includegraphics[width=17.6cm, height = 6cm]{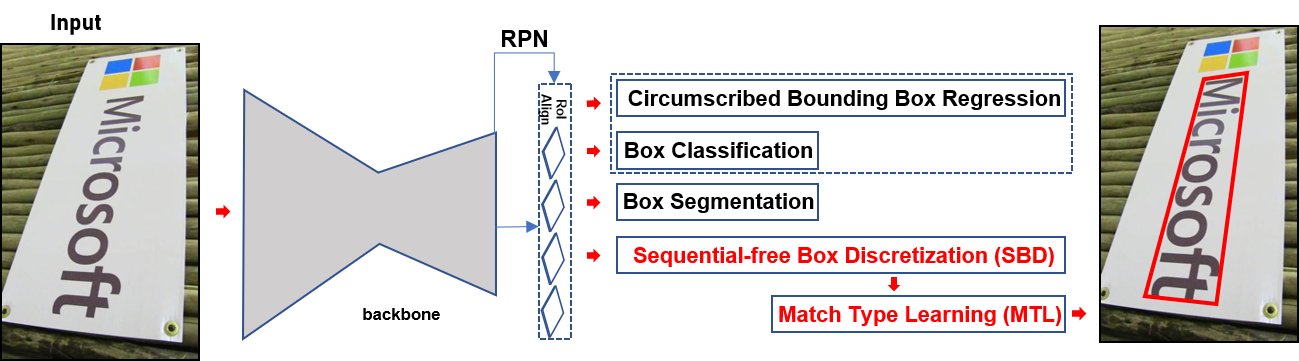}}
  \caption{Overall framework. SBD is connected to the Mask R-CNN as an additional branch. The backbone is ResNet-50-FPN in this paper.}\label{fig:overall}
\end{figure*}

Motivated by this issue, this paper proposes a simple but effective method called Sequential-free Box Discretization (SBD) that can parameterize the bounding boxes into key edges. Basically, to avoid LC issue, the basic idea is to find at least four invariant points (e.g., mean center point, and intersecting point of the diagonals) that are irrelevant to the label sequence and we can use these invariant points to inversely deduce the bounding box coordinates. To simplify parameterization, a novel module called key edge (KE) is proposed to learn the bounding box.

Experiments on many public scene text benchmarks, including MLT \cite{nayef2017icdar2017}, MSRA-TD500 \cite{Yao2012Detecting}, and ICDAR 2015 Robust Reading Competition Challenge 4 ``Incidental scene text localization'' \cite{karatzas2015icdar}, all demonstrated that our method can outperform previous state-of-the-art methods in terms of Hmean. Moreover, ablation studies showed that by seamlessly integrating SBD in Mask R-CNN framework, the detection result can be substantially improved.
On multi-oriented ship detection dataset HRSC2016 \cite{liu2017rotated}, our method can still perform the best, further showing its promising generalization ability.

The main contributions of this paper are manifold: 1) We propose an effective SBD method which can not only solve LC issue but also improve the omnidirectional text detection performance; 2) SBD and its derived post-processing methods can further guarantee tighter and more accurate detections; 3) our method can substantially improve Mask R-CNN and achieve the state-of-the-art performance on various benchmarks.

\section{Related Work}
The mainstream multi-oriented scene text detection methods can be roughly divided into segmentation-based methods and non-segmentation-based methods.

\subsection{Segmentation-based Method}

Most of segmentation-based text detection methods are mainly built and improved from the FCN \cite{long2015fully} or Mask R-CNN \cite{he2017mask}. Segmentation-based methods are not SLS methods because the key of segmentation-based method is to conduct pixel-level classification. However, how to accurately separate the adjacent text instances is always a tough issue for segmentation-based methods. Recently, many methods are proposed to solve this issue. For examples, PixelLink \cite{deng2018pixellink} additionally learns 8-direction information for each pixel to highlight the text margin; \cite{lyu2018multi} proposes a corner detection method to produce position-sensitive score map; and \cite{wu2017self} defines text border map for effectively distinguishing the instances. 

\subsection{Non-segmentation-based Method} 
Segmentation-based methods require or post-processing steps to group the positive pixels into final detection results, which may easily be affected by the false positive pixels. Non-segmentation methods can directly learn the exact bounding box to localize the text instances. For examples, \cite{liao2018rotation} predicts text location by using different scaled feature; \cite{liu2017deep} and \cite{ma2018arbitrary} utilize quadrilateral and rotated anchors to detect the multi-oriented text; \cite{liao2018textboxes++} utilizes carefully-designed anchors to localize text instances; \cite{zhou2017east} and \cite{he2017deep} directly regress the text sides or vertexes of the text instances. Although non-segmentation methods can also achieve superior performance, most of the non-segmentation methods are SLS methods, and thus they might easily be affected by the label sequence.

\section{Methodology}
In this section, we describe the details of the SBD. SBD is theoretically suitable for any general object detection framework, but in this paper we only build and validate SBD on Mask R-CNN. The overall framework is illustrated in Figure~\ref{fig:overall}.

\subsection{Sequential-free Box Discretization}
The main goal of omnidirectional scene text detection is to accurately predict the compact bounding box which can be rectangular or quadrilateral. As introduced in Section \ref{sec:intro}, introducing quadrilateral bounding box can also bring the LC issue. Therefore, instead of predicting label-sensitive distances or coordinates, SBD discretizes the quadrilateral GT box into 8 lines that only contain invariant points, which are called key edges (KE). As shown in Figure \ref{fig:SBD}, eight KEs in this paper are discretized from the original coordinates: minimum $x$ ($x_{min}$) and $y$ ($y_{min}$); the second smallest $x$ ($x_2$) and $y$ ($y_2$); the second largest $x$ ($x_3$) and $y$ ($y_3$); maximum $x$ ($x_{max}$) and $y$ ($y_{max}$). 

As shown in the Figure \ref{fig:overall} and \ref{fig:SBD}, the inputs of SBD are the proposals processed by RoIAlign \cite{he2017mask}; the feature map is then connected to stacked convolution layers and then upsampled by $2\times$ bilinear upscaling layers, and the resolution of output feature maps $F_{out}$ from deconvolution is restricted to $M\times M$. For each of the x-KEs and y-KEs, we use $1\times M$ and $M\times 1$ convolution kernels with four output channels to shrink the transverse and longitudinal features, respectively; the number of the output channels are set to the same as the number of x-KEs or y-KEs, respectively. After that, we assign corresponding positions of the GT KEs to each output channel and update the network by minimizing the cross-entropy loss {$L_{KE}$} over a M-way softmax output. We found detection in such classification manner instead of regression would be much more accurate. 

Taking $t_{i}$ ($t$ can be $x$ or $y$, and $i$ can be min, 2, 3, max) as an example, we do not directly learn the $t_{i}$-th KE; instead, the GT KE is the vertical line $t_{ihalf}$, and $t_{ihalf} = (t_{i} + t_{mean})/2$, where $t_{mean}$ represents the $t$ value of the mean central point of the GT box. Learning $t_{ihalf}$ has two important advantages:
\begin{figure}[!t]
  \centering
  \centerline{\includegraphics[width=8.6cm, height = 5.5cm]{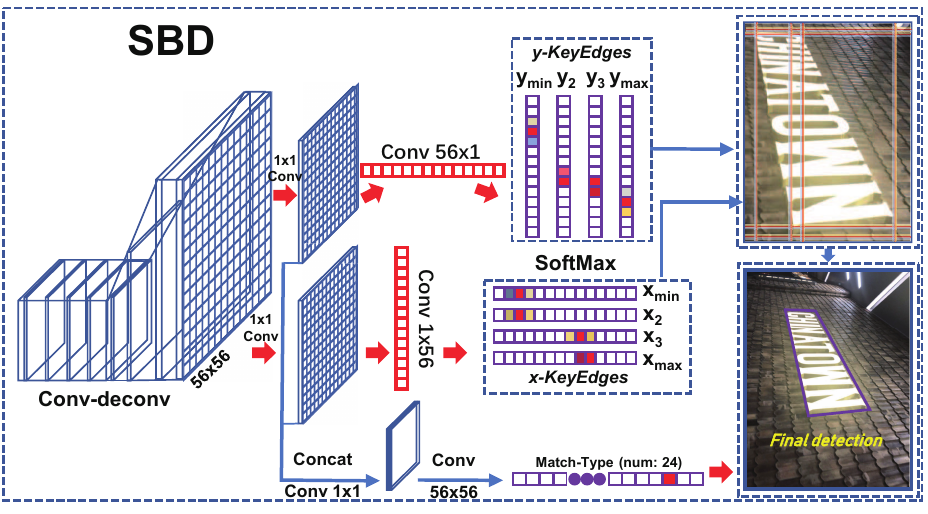}}
  \caption{Illustration of SBD. The resolution $M$ in this paper is simply set to 56.}\label{fig:SBD}
\end{figure}
\begin{itemize}
  \item Breaking RoI restriction. The original Mask R-CNN only learns to predict inside the RoI, and if parts of the target instances are outside the RoI, it would be impossible to recall these missing pixels. However, as shown in Figure \ref{fig:break}, learning $t_{ihalf}$ can output the real border even if the border is outside the RoI.
  \begin{figure}[!t]
    \centering
    \centerline{\includegraphics[width=8.6cm, height = 3.5cm]{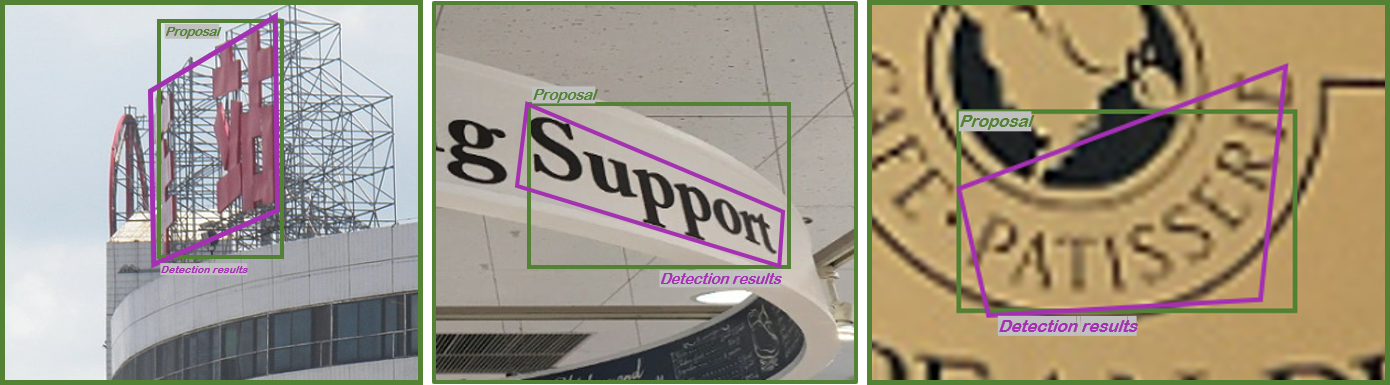}}
    \caption{Detection examples that the results of SBD can break the restriction of proposal (RoI).}\label{fig:break}
  \end{figure}
  \item Even if the border of the text instance is outside the RoI, in most cases, the $t_{ihalf}$ remains inside the RoI. Therefore, the integration of the text instance can be guaranteed and loss can be well propagated (because if a learning target is outside the RoI, the loss is zero).
\end{itemize}

Formally, a multi-task loss on each foreground RoI is defined as $L = L_{cls} + L_{box} + L_{mask} + L_{ke}$. The first three terms $L_{cls}$, $L_{box}$, and $L_{mask}$ are the same as \cite{he2017mask}. 
It is worth mentioning that \cite{he2017mask} pointed out that the additional keypoint branch reduces the performance of box detection in Table 5; however, from our experiments, the proposed SBD is the key component for boosting detection performance, which we think is mainly because: 1) For keypoint learning, there are $M^{2}$ classes against each other, while for SBD, the number of competitive pixels is only $M$; 2) the keypoint might not be very explicit for a specific point (it could be a small region), while the KEs produced by SBD represent the borders of GT instances, which are absolute and exclusive, and thus the supervision information would not be confused.

\subsubsection{Match-Type Learning}
Based on the box discretization, we can learn the values of all $x$ and $y$, but we do not know which y-KEs should be matched to which x-KEs. Intuitively, as shown in the top right of the Figure \ref{fig:SBD}, designing a proper matching procedure is a very important issue, otherwise the detection results could be significantly worse.

To solve this problem, we propose a simple but effective match-type learning (MTL) method. As shown in Figure \ref{fig:SBD}, we concatenate the x-KE and y-KE feature maps followed by $1\times 1$, $M\times M$ convolutions, and softmax loss is used to learn a total of 24 ($A_{4}^{4}$) match-types (because we have 4 x-KEs and y-KEs), including \{1234,1243,1324, ..., 4312, 4321\}. For example, in the case of the Figure \ref{fig:overall}, the predicted match-type is ``2413'' which represents the matching results are ($x_{min}$, $y_{2}$), ($x_{2}$, $y_{max}$), ($x_{3}$, $y_{min}$), ($x_{max}$, $y_{3}$). 

\begin{figure}[!t]
  \centering
  \centerline{\includegraphics[width=8.6cm, height = 6cm]{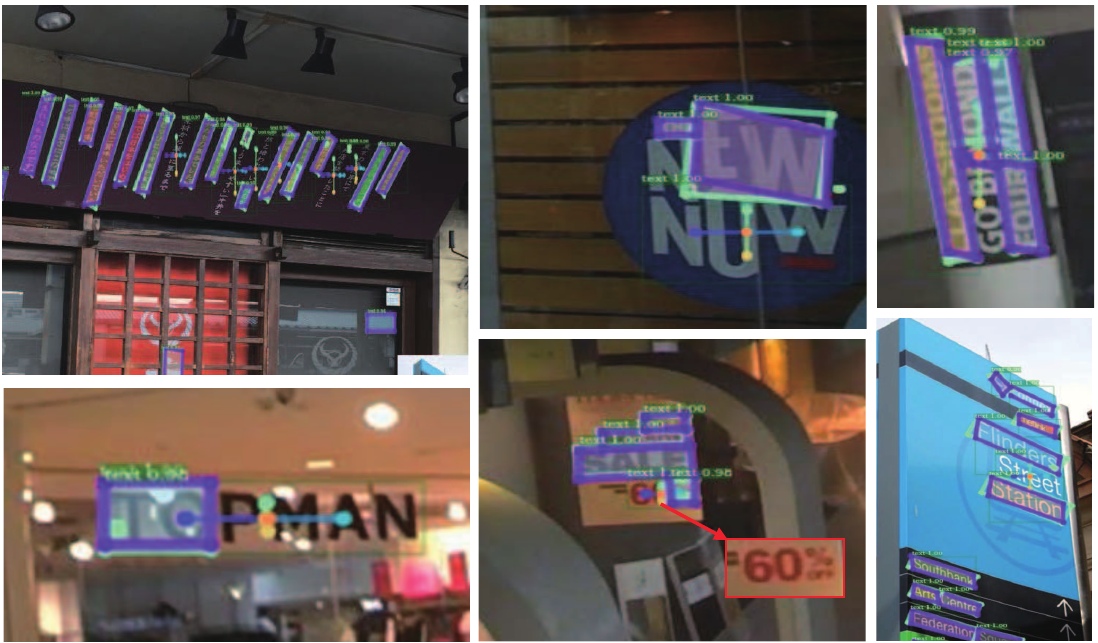}}
  \caption{Examples that SBD can recall many instances that segmentation branch fails to recall. Green bounding boxes and scores represent RoIs. Rotated cyan bounding box and transparent pixels represent the result from segmentation branch. Transparent pixels are predicted by mask branch. Purple quadrangles are final detection results from SBD. KEs are simplify by colorful points.}\label{fig:maskMiss}
\end{figure}

During training, we find the MTL can be very easy to learn and the loss can quickly converge within ten thousand iterations with 1 image per batch. Moreover, in some cases, the segmentation branch would somehow produce non-positive pixel while both the proposals and SBD predictions are accurate, as shown in Figure \ref{fig:maskMiss}. Through MTL, SBD can output the final bounding box and improve the detection performance by offsetting the weakness of segmentation branch. 

\subsubsection{Rescoring and Post Processing}

Based on our observations, some unconsolidated detections could also have virtual high confidence. This is mainly because the confidence outputted from the softmax in Fast R-CNN \cite{girshick2015fast} is a classification loss but not for localization. Therefore, the compactness of the bounding box cannot be directly supervised by the score.

\begin{figure}[!t]
  \centering
  \centerline{\includegraphics[width=8.6cm, height = 5cm]{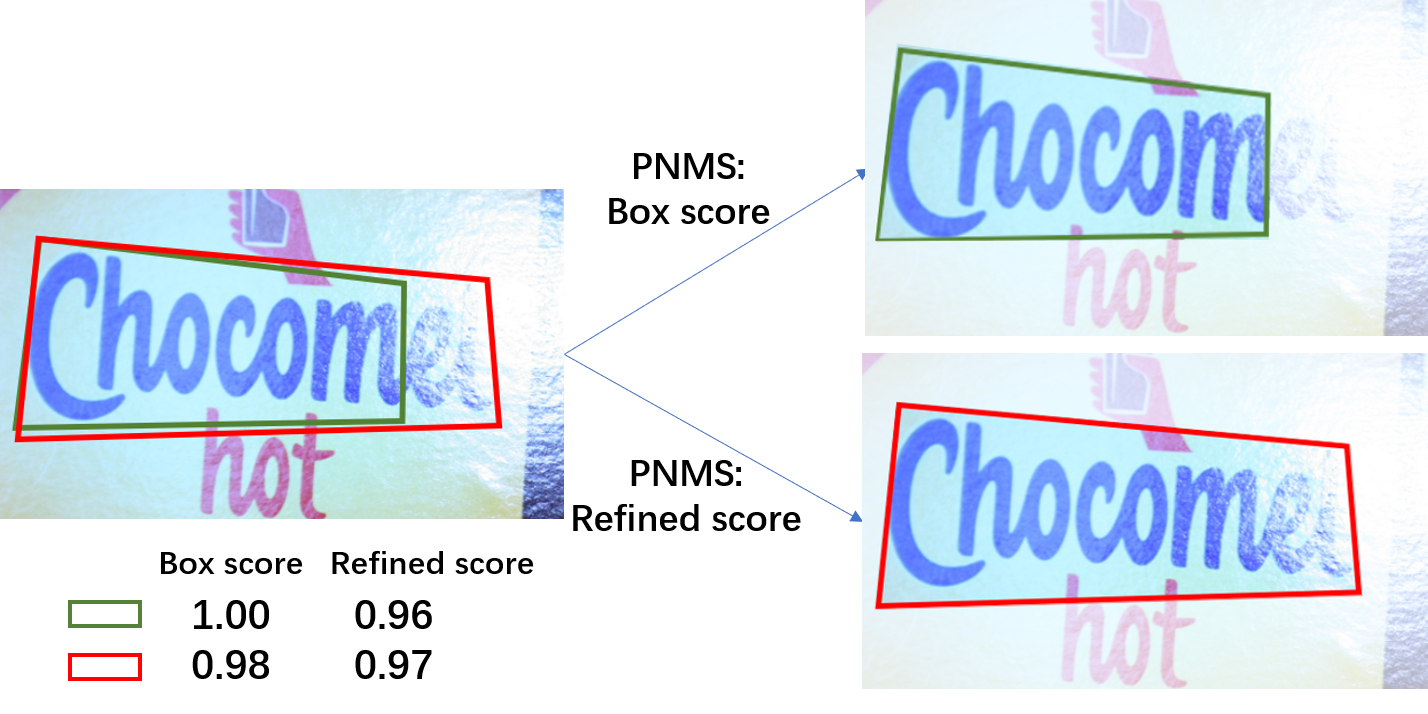}}
  \caption{Example of the effect of rescoring. Original confidence is mainly for classification, while our refined score further considers the localization possibility.}\label{fig:rescore}
\end{figure}

We thus compute a refined confidence that takes the advantages of SBD prediction which learns the specific position of the final detections. Formally, we refine final instance-level detection score as follow:
\begin{equation}\label{eq:rescore}
  score(\Re) = \frac{(2-\gamma)S_{box} + \gamma S_{SBD}}{2},
\end{equation}
where, $\gamma$ is the weighting coefficient, and it satisfies $0\leq\gamma$, and $\gamma\leq2$. $S_{box}$ is the original softmax confidence for the bounding box, and $S_{SBD}$ represents the mean score of all the KEs, which is defined below: 
\begin{equation}\label{eq:sKE}
  S_{SBD} = \frac{1}{K}\sum_{k=1}^{K}\max_{x_{i}}f_{k}(x_{i}),
\end{equation}
where, $K$ is the number of the KEs (which is 8, including 4 x-KEs and 4 y-KEs); $f$ is the function to calculate the sum of adjacent 5 scores. We have found that using Equation (\ref{eq:rescore}) can not only suppress some false positives but also make the results more reliable. Examples are shown in Figure \ref{fig:rescore}.

\section{Experiments}

\subsection{Implemented Details}
We used synthetic data \cite{Gupta2016Synthetic} to pretrain the model and finetuned on the provided training data from MLT \cite{nayef2017icdar2017}, and ICDAR 2015 \cite{karatzas2015icdar}. For MSRA-TD500 \cite{Yao2012Detecting}, because the limited number of the Chinese samples, we pretrained the model from 4k well annotated samples from \cite{Shi2017ICDAR2017} and finetuned by official training samples. 

The number of maximum iterations is 40 epochs for each dataset on four NVIDIA 1080ti GPUs. The initial learning rate is $10^{-2}$ and reduces to $10^{-3}$ and $10^{-4}$ on the 25th and 32th epoch, respectively. In order to balance the learning weights of all branches, the weights of KEs and match-type learning are empirically restricted to 0.2 and 0.01, respectively.

The resolutions of training images were randomly selected from 600 to 920 with the interval of 40, and the maximum size was restricted to 1480.
For testing, we only used single scale for all datasets (public methods have numerous settings for multi-scale testing, which is hard to conduct a fair comparison), and the scale and maximum size is (1200, 1600). Polygon non-maximum suppression (PNMS) \cite{yuliang2017detecting} with threshold 0.2 is used to suppress the redundant detections.

\subsection{Experiments on the Scene Text Benchmarks}

\begin{figure}[!t]
  \centering
  \centerline{\includegraphics[width=8.6cm, height = 5cm]{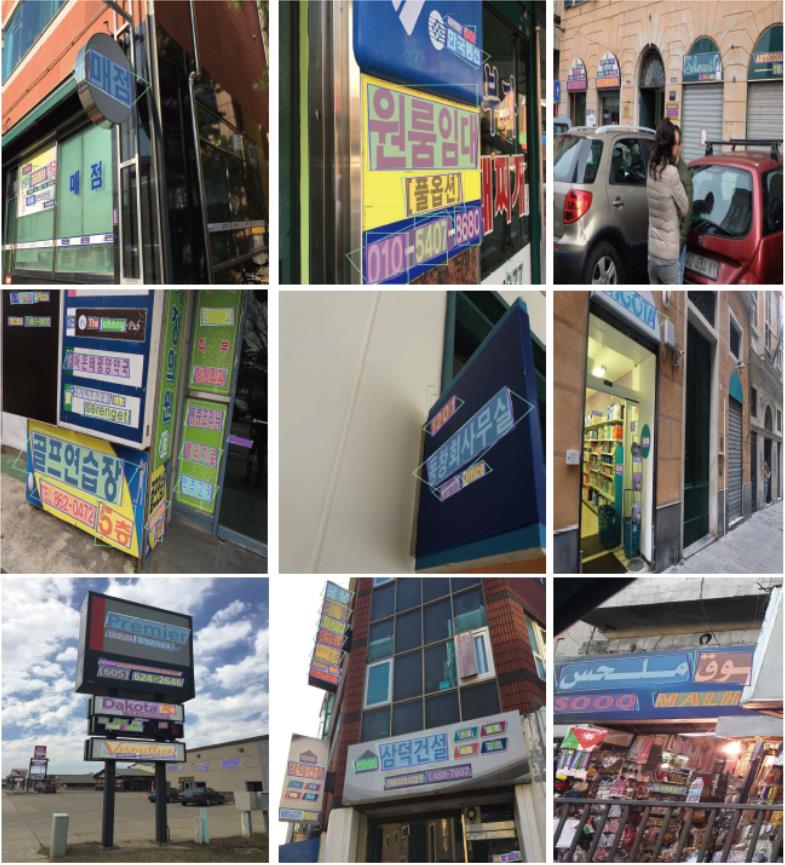}}
  \caption{Examples of detection results. Purple detections are the final detection results of SBD. The transparent regions are the segmentation results from the segmentation branch, and rotated rectangles are the minimum area bounding boxes grouped by the transparent regions. Horizontal thin green bounding boxes are the rois. Zoom in for better visualization.}\label{fig:dettext}
\end{figure}

\paragraph{ICDAR 2017 MLT.} \cite{nayef2017icdar2017} is the largest multi-lingual (9 languages) oriented scene text dataset, including 7.2k training samples, 1.8k validation samples and 9k testing samples. The challenges of this dataset are manifold: 1) Different languages have different annotating styles, e.g., most of the Chinese annotations are long (there is not specific word interval for a Chinese sentence) while most of the English annotations are short, the annotations of Bangla or Arabic may be frequently entwined with each other; 2) more multi-oriented, perspective distortion text on various complexed backgrounds; 3) many images have more than 50 text instances. All instances are well annotated with compact quadrangles. The results of MLT are given in Table \ref{tab:mlt}. Our method outperforms previous state-of-the-art methods by a large margin, especially in terms of recall rate. Some of the detection results are visualized in Figure \ref{fig:dettext}. Instead of merely using segmentation predictions to group the rotated rectangular bounding boxes, SBD can directly predict the compact quadrilateral bounding boxes which should be more reasonable. Although there are some text instances missed, most of the text can be robustly recalled.

\paragraph{MSRA-TD500.} \cite{Yao2012Detecting} is a text-line based oriented dataset with 300 training images and 200 testing images captured from indoor and outdoor scenes. Although this dataset contains less text per image and most of the text is clean, the major challenge of this dataset is that most of the text in this dataset has the large variance in orientations. The results of MSRA-TD500 are given in Table \ref{tab:msra}. Although our method is slower than some of the previous methods, it has a significant improvement in terms of the $Hmean$, which demonstrates its robustness in detecting long and strong tilted instances.

\paragraph{ICDAR 2015 Incidental Scene Text.} \cite{karatzas2015icdar} is one of the most popular benchmarks for oriented scene text detection. The images are incidentally captured mainly from streets and shopping malls, and thus the challenges of this dataset rely on the oriented, small, and low resolution text. This dataset contains 1k training samples and 500 testing samples, with about 2k content-recognizable quadrilateral word-level bounding boxes. The results of ICDAR 2015 are given in Table \ref{tab:ic15}. From this table, we can observe that our method can still perform the best. 

\begin{table}[!t]
\centering
\small
\begin{tabular}{lccc}
  \hline
  Algorithms  & $R (\%)$  & $P (\%)$ & $H (\%)$ \\
  \hline
  \cite{nayef2017icdar2017} & 25.59 & 44.48 & 32.49 \\
  \cite{nayef2017icdar2017} & 34.78 & 67.75 & 45.97 \\
  \cite{ma2018arbitrary} & 67.0 & 55.0 & 61.0 \\
  \cite{ma2018arbitrary} & 55.5 & 71.17 & 62.37 \\
  \cite{nayef2017icdar2017} & 69.0 & 67.75 & 45.97 \\
  \cite{nayef2017icdar2017} & 62.3 & 80.28 & 64.96 \\
  \cite{zhong2018anchor} & 66.0 & 75.0 & 70.0 \\
  \cite{lyu2018multi} (SS) & 55.6 & \bf 83.8 & 66.8 \\
  \cite{liu2018fots} (SS) & 62.3 & 81.86 & 70.75 \\
  \hline
  {\bf Proposed method} & \bf 70.1 & 83.6 & \bf 76.3 \\
  \hline
\end{tabular}
\caption{Experimental results on MLT dataset. SS represents single scale. R: Recall rate. P: Precision. H: Harmonic mean of R and P. Note that we only use single scale for all experiments. }
\label{tab:mlt}
\end{table}

\begin{table}[!t]
\centering
\small
\begin{tabular}{lcccc}
  \hline
  Algorithms  & $R (\%)$  & $P (\%)$ & $H (\%)$  & FPS\\
  \hline
  \cite{kang2014orientation} & 62.0 & 71.0 & 66.0 & - \\
  \cite{zhang2016multi} & 67.0 & 83.0 & 74.0 & 0.48 \\
  \cite{yao2016scene} & 75.3 & 76.5 & 75.9 & 1.61 \\
  \cite{zhou2017east} & 67.4 & 87.3 & 76.1 & {\bf 13.2} \\
  \cite{shi2017detecting} & 70.0 & 86.0 & 77.0 & 8.9 \\
  \cite{he2017deep} & 70.0 & 77.0 & 74.0 & 1.1 \\
  \cite{wu2017self} & 78.0 & 77.0 & 77.0 & - \\
  \cite{deng2018pixellink} & 73.2 & 83.0 & 77.8 & - \\
  \cite{lyu2018multi} & 76.5 & 87.6 & 81.5 & 5.7 \\
  \cite{liao2018rotation} & 73.0 & 87.0 & 79.0 & 10 \\
  \hline
  {\bf Proposed method} & {\bf 80.5} & \bf 89.6 & \bf 84.8 & 3.2\\
  \hline
\end{tabular}
\caption{Experimental results on MSRA-TD500 benchmark.}
\label{tab:msra}
\end{table}

\begin{table}[!t]
\centering
\small
\begin{tabular}{lccc}
  \hline
  Algorithms  & $R (\%)$  & $P (\%)$ & $H (\%)$ \\
  \hline
  \cite{zhang2016multi} & 43.0 & 71.0 & 54.0\\
  \cite{tian2016detecting} & 52.0 & 74.0 & 61.0\\
  \cite{shi2017detecting} & 76.8  & 73.1 & 75.0\\
  \cite{liu2017deep} & 68.2 & 73.2 & 70.6\\
  \cite{zhou2017east} & 73.5 & 83.6 & 78.2\\
  \cite{hu2017wordsup} & 77.0 & 79.3 & 78.2\\
  \cite{liao2018rotation} & 79.0 & 85.6 & 82.2\\
  \cite{deng2018pixellink} & 82.0 & 85.5 & 83.7\\
  \cite{ma2018arbitrary} & 82.2 & 73.2 & 77.4\\
  \cite{lyu2018multi} & 79.7 & {\bf 89.5} & 84.3\\
  \cite{he2017deep} & 80.0 & 82.0 & 81.0 \\
  \hline
  {\bf Proposed method} & {\bf 83.8} & 89.4 & {\bf 86.5}\\
  \hline
\end{tabular}
\caption{Experimental results on ICDAR 2015 dataset. For fair comparison, this table only listed the single scale results without recognition supervision.}
\label{tab:ic15}
\end{table}

\begin{figure}[htb]
  \centering
  \centerline{\includegraphics[width=8.6cm, height = 5.5cm]{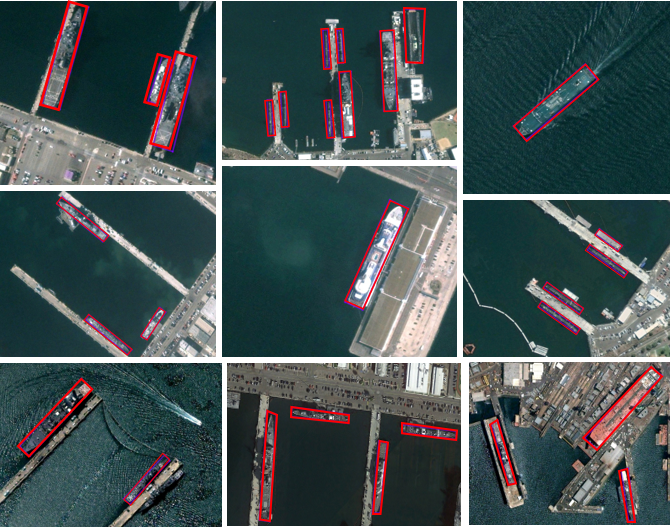}}
  \caption{ Experimental results on HSRC 2016. The detections are highlighted with red bounding boxes. }\label{fig:shipdet}
\end{figure}

\subsection {Ablation Studies}
\begin{figure}[!t]
  \centering
  \centerline{\includegraphics[width=8.0cm, height = 5cm]{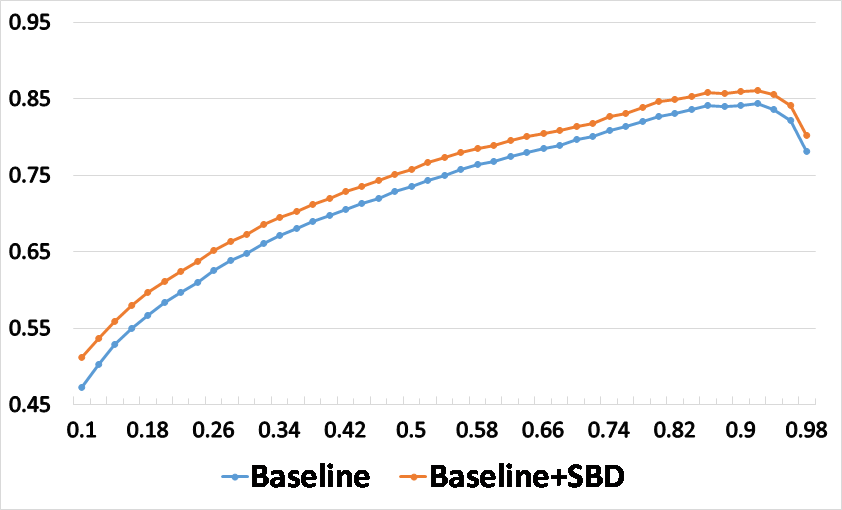}}
  \caption{Ablation study on ICDAR 2015 benchmark. X-axis represents confidence threshold and Y-axis represents Hmean result. Baseline represents Mask R-CNN. By integrating with proposed SBD, the detection results can be substantially better than the results of Mask R-CNN baseline.}\label{fig:ablation}
\end{figure}

In this section, we further conducted ablation studies to validate the effectiveness of SBD, and the results are shown in Table \ref{tab:ablat} and Figure \ref{fig:ablation}. Table \ref{tab:ablat} showed that adding SBD can lead to 2.4\% improvement in terms of Hmean. One reason is that the SBD can recall more instances, as discussed in Section 3 and shown in Figure \ref{fig:maskMiss}; the other reason maybe the SBD branch can bring the effect of mutual promotion just like how segmentation branch improves the performance of the Mask R-CNN. In addition, Figure \ref{fig:ablation} showed our method can substantially outperform the baseline Mask R-CNN under different confidence thresholds of the detections, which further demonstrated its effectiveness.

We also conducted experiments to compare and validate different methods' ability of resistant to the LC issue. Specifically, we first trained the East \cite{zhou2017east}, CTD \cite{yuliang2017detecting}, and proposed method with original 1k training images of ICDAR 2015 dataset. Then, we randomly rotated the training images among $[0^\circ, 15^\circ, 30^\circ, ..., 360^\circ]$ and randomly picked up additional 2k images from the rotated dataset to finetune on the these three methods. The results are given in Table \ref{tab:sequence}, which demonstrated the powerful sequential-free ability of the proposed SBD.

\begin{table}[!t]
\centering
\small
% \small
\begin{tabular}{c|c|c|c|c}
  \hline
   & Textboxes++ & East  & CTD  & Ours \\
  \hline
  $\Delta$ Hmean & \bf \textcolor[RGB]{0,160,0}{$\downarrow$ 9.7\%} & \bf \textcolor[RGB]{0,160,0}{$\downarrow$ 13.7\%} & \bf \textcolor[RGB]{0,160,0}{$\downarrow$ 24.6\%} & \bf \color{red}{$\uparrow$ 0.3\%} \\
  \hline
\end{tabular}
\caption{Comparison on ICDAR 2015 dataset showing different methods' ability of resistant to the LC issue (by adding rotated pseudo samples). East and CTD are both SLS methods.}
\label{tab:sequence}
\end{table}

\begin{table}[!t]
\centering
\newcommand{\tabincell}[2]{\begin{tabular}{@{}#1@{}}#2\end{tabular}}
\small
% \small
\begin{tabular}{c|c|c}
  \hline
  Datasets & Algorithms & Hmean \\
  \hline
  \multirow{3}*{\bf ICDAR2015} & Mask R-CNN baseline & 83.5\% \\
                          \cline{2-3}
                          & Baseline + {\color{red} SBD} & 85.9\% (\bf \color{red} $\uparrow$ 2.4\%) \\
                          \cline{2-3}
                          & Baseline + SBD + {\color{red}Rescoring} & 86.5\% (\bf \color{red} $\uparrow$ 0.6\%) \\
  \hline
\end{tabular}
\caption{Ablation studies to show the effectiveness of the proposed method. The $\gamma$ of rescoring is set to 1.4 (best practice).}
\label{tab:ablat}
\end{table}
 
\subsection{Experiments on the Ship Detection Benchmark}
To demonstrate generalization ability of SBD, we further evaluated and compared SBD on Level 1 task of the HRSC2016 dataset \cite{liu2017rotated} to show our method's performance on multi-directional object detection. The ship instances in this dataset might appear in various orientations, and annotating bounding box is based on rotated rectangles. There are 436, 181, and 444 images for training, validating, and testing set, respectively. The evaluating metric is the same as \cite{karatzas2015icdar}. Only the training and validation sets are used for training, and because of the small amount of the training data, the whole training procedure took us only about two hours.

The result showed that our method can easily surpass previous methods by a large margin (as shown in Table \ref{tab:hrsc}), $7.7\%$ higher than recent state-of-the-art RRD \cite{liao2018rotation} in mAP score. Some of the detection results are presented in Figure \ref{fig:shipdet}. Both the quantitative and qualitative results all show that the proposed method can perform well on common oriented object detections even with very limited training data, further demonstrating its powerful generalization ability.

\begin{table}[!t]
\centering
\small
% \small
\begin{tabular}{c|cccc}
  \hline
  Algorithms  & mAP \\
  \hline
  \cite{girshick2015fast,liao2018rotation} & 55.7\\
  \cite{girshick2015fast,liao2018rotation} & 69.6 \\
  \cite{girshick2015fast,liao2018rotation} & 75.7\\
  \cite{liao2018rotation} & 84.3 \\
  \hline
  {\bf Proposed method} &  {\bf 93.7} \\
  \hline
\end{tabular}
\caption{Experimental results on HRS\_2016 dataset. }
\label{tab:hrsc}
\end{table}

\section{Conclusion}
This paper proposed SBD - a novel method that uses discretization methodology for oriented scene text detection.

SBD solves the LC issue by discretizing the point-wise prediction into sequential-free KEs that only contain invariant points, and using a novel match-type learning method to guide the compound mode. Benefiting from SBD, we can improve the reliability of the confidence of the bounding box and adopt more effective post-processing methods to improve performance.

Experiments on various oriented scene text benchmarks (MLT, ICDAR 2015, MSRA-TD500) all demonstrate the outstanding performance of the SBD. To test generalization ability, we further conducted an experiment on oriented general object dataset HRSC2016, and the results showed that our method can outperform recent state-of-the-art methods with a large margin.

\section*{Acknowledgements}
This research is supported in part by the National Key Research and Development Program  of China (No. 2016YFB1001405), GD-NSF (no.2017A030312006),  NSFC (Grant No.: 61673182, 61771199), and GDSTP (Grant No.:2017A010101027) , GZSTP(no. 201704020134).

\bibliographystyle{named}
\bibliography{IJCAI-BDN}

\begin{figure*}[!t]
  \centering
  \centerline{\includegraphics[width=17.6cm, height = 8cm]{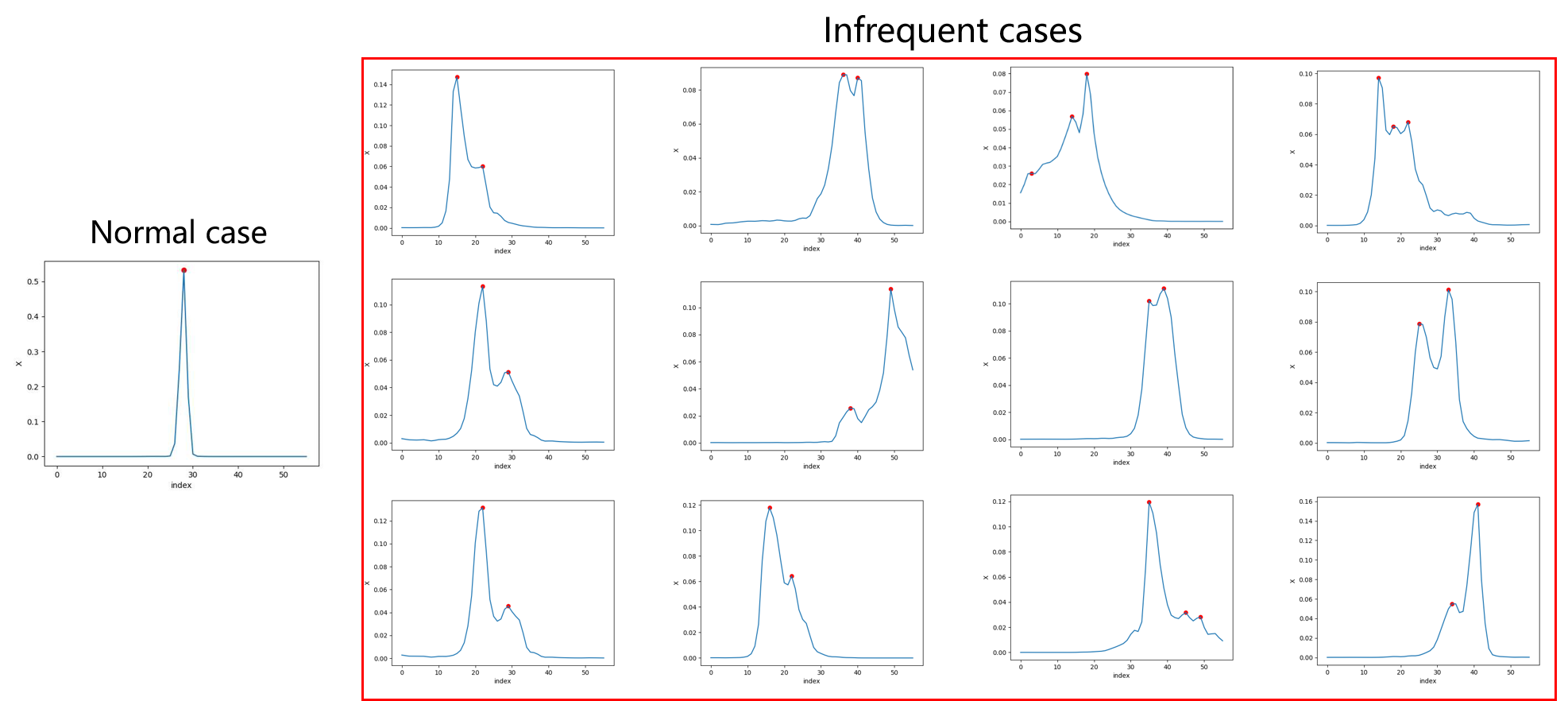}}
  \caption{Examples of KE score results.}\label{fig:kescores}
\end{figure*}
\newpage
\section*{Appendix}

Some additional data and figures are provided here for better understanding our method. Our method is built on MaskRCNN-benchmark\footnote{https://github.com/facebookresearch/maskrcnn-benchmark}, which is based on pytorch framework.

\paragraph{KE score.} Figure \ref{fig:kescores} shows some example results of the ke scores. Normally, detection results will produce the similar shapes as the normal case in Figure \ref{fig:kescores}. Note that even if in the normal case, the highest score may still obviously below 1.0, and that explains why we use the sum of adjacent 5 score in the rescoring operation.

\begin{figure}[!t]
  \centering
  \centerline{\includegraphics[width=8.6cm, height = 4.5cm]{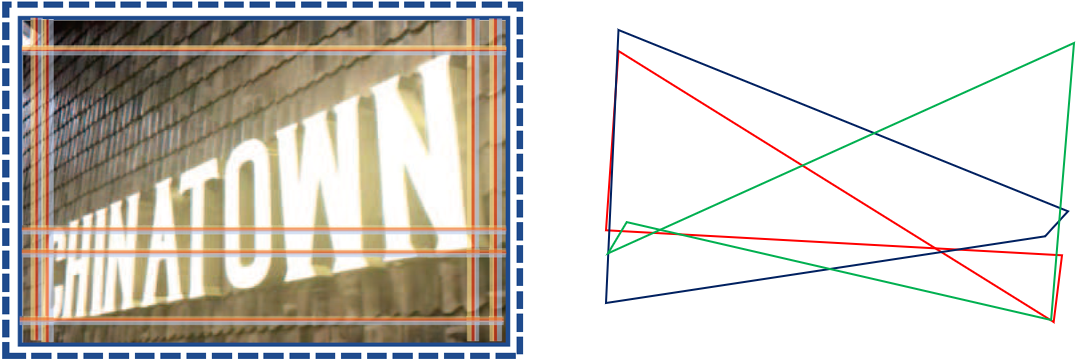}}
  \caption{Examples of wrong match type results (different colors).}\label{fig:wrong_mtl}
\end{figure}

\begin{figure}[!t]
  \centering
  \centerline{\includegraphics[width=8.6cm, height = 6.5cm]{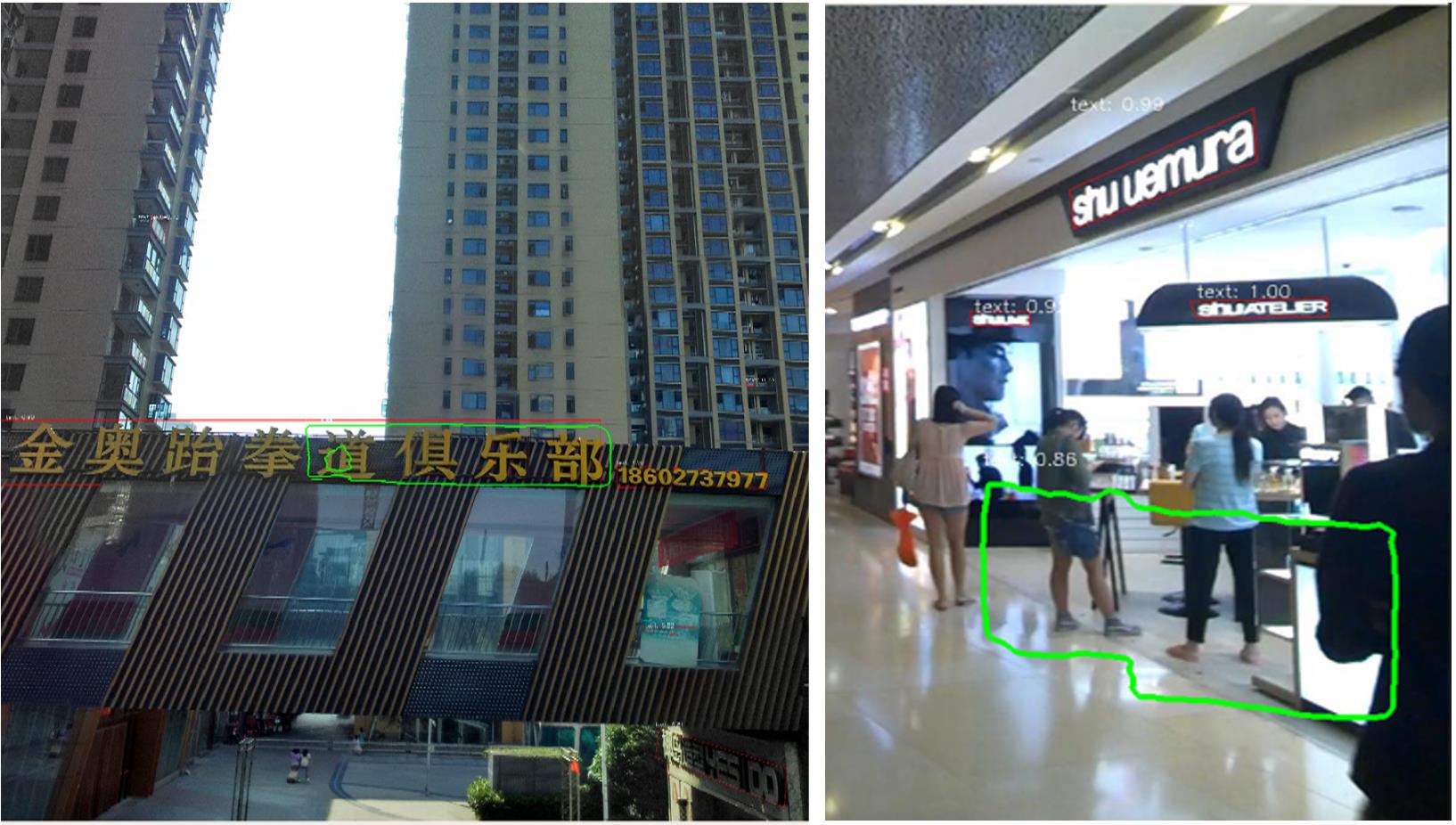}}
  \caption{Examples of false positives suppression.}\label{fig:fpsuppress}
\end{figure}

\paragraph{False positive suppression.} Match type can also be used for false positives. Because for some false positives, there is not clear edge, and in such case the match type learning may predict an abnormal result as shown in Figure \ref{fig:wrong_mtl}. These abnormal results can be easily removed by judging if the quadrangle is valid (sides should only have two intersections on the head and tail). By doing so, we can further eliminate some false positives that might cheat mask branch, as shown in Figure \ref{fig:fpsuppress}.

\paragraph{OKS-NMS.} [Papandreou \textit{et al.}, 2017] adopted object keypoint similarity non-maximum suppression (NMS-OKS) that can be effective to suppress some unnecessary box-in-box. We can follow similar implement on our KEs detections except removing $\sigma^{2}$, which is because all KEs should weight the same. The formula is given as follows:
\begin{equation}\label{eq:oes}
  OKS_{p}= \frac{\sum_{i}exp\{-d_{pi}^{2}/2s_{p}^{2}\}\delta(v_{pi}-1)}{\sum_{i}\delta(v_{pi} = 1)}.
\end{equation}

\end{document}